\documentclass[11pt]{article}

\usepackage[final]{acl}

\usepackage{times}
\usepackage{latexsym}

\usepackage[T1]{fontenc}

\usepackage[utf8]{inputenc}

\usepackage{microtype}

\usepackage{inconsolata}

\usepackage{graphicx}

\usepackage{algorithm}
\usepackage{algorithmic}
\usepackage{booktabs}
\usepackage{amsmath}
\usepackage{amssymb}
\usepackage{pifont}
\newcommand{\xmark}{\ding{55}}%
\usepackage{CJKutf8}
\newcommand{\zh}[1]{\begin{CJK}{UTF8}{gbsn}#1\end{CJK}}
\title{Vision-Braille: A Curriculum Learning Toolkit and\\Braille–Chinese Corpus for Braille Translation}

\author{
    \textbf{Alan Wu\textsuperscript{2}$^*$},
    \textbf{Ye Yuan\textsuperscript{1}},
    \textbf{Zhiping Xiao\textsuperscript{3}},
    \textbf{Ming Zhang\textsuperscript{1}},
\\
\\
    \textsuperscript{1}State Key Laboratory for Multimedia Information Processing, School of Computer Science, \\ PKU-Anker LLM Lab, Peking University \\
    \textsuperscript{2}College of Agriculture and Life Sciences, Cornell University \\
    \textsuperscript{3}Computer Science Department, University of California at Los Angeles
\\
    \small{
        \textbf{Correspondence:} \href{mailto:mzhang_cs@pku.edu.cn}{Ming Zhang (\texttt{mzhang\_cs@pku.edu.cn})}
    }
}

\newif\ifshowcomment
\showcommenttrue

\ifshowcomment
\newcommand{\yuanye}[1]{\textcolor{orange}{[Yuanye: #1]}}
\else
\newcommand{\yuanye}[1]{}
\fi

\newcommand{\methodname}{Curriculum-Braille}

\newcommand{\stageone}{Foundation}
\newcommand{\stagetwo}{Long-range}
\newcommand{\stagethree}{Tone curriculum}
\newcommand{\stagefour}{Consolidation}

\begin{document}
\maketitle
\def\thefootnote{*}\footnotetext{Work done during the remote internship at Dlib Group, School of Computer Science, Peking University.}
\def\thefootnote{\arabic{footnote}}

\begin{abstract}
We present Vision-Braille, the first publicly available end-to-end system for translating Chinese Braille extracted from images into written Chinese. This system addresses the unique challenges of limited annotated resources and tone omission. It integrates a robust Braille OCR pipeline with an LLM fine-tuned for sequence-to-sequence translation. We construct a synthetic Braille-Chinese corpus, including tone-omission variants that mimic authentic Braille writing habits. We fine-tune the model using a four-stage curriculum: starting with sentence-level data with full tone markers, progressing to passage-level data, then applying a tone-omission schedule of decreasing retention, and finally consolidating on passages with heavy tone omission. On passage-level translation with 10\% tone retention, \methodname{} achieves 83.28 BLEU. Vision-Braille offers an inclusive NLP solution that empowers students with visual impairments to participate in mainstream education by enabling teachers to grade Braille homework without extensive training. Our code and data are available at \url{https://anonymous.4open.science/r/EMNLP_2026_Supp_Code_Data-2F6D}.
\end{abstract}

\section{Introduction}
    Among the $4$ million registered people with visual disabilities in China~\cite{china_disabled_persons_federation_chinese_2022}, there is a special group of $150,000$ school-age blind children~\cite{educational_issue_braille} who face the risk of ``double-blind'' (physically disabled and illiterate). We have found that even though the population is staggering, education resources are scarce: only 10 percent can attend one of the $11$ Senior High Schools for Blind Students nationwide, located in major cities~\cite{china_disabled_persons_federation_chinese_2022}. Although the Chinese Bureau of Education enacted the specialized form of GaoKao, allowing participants to read Braille exam papers and write in Braille, this group was almost absent in higher education: only $15$ students nationwide participated in this special GaoKao. This number was small compared with the total number of $13$ million GaoKao students~\cite{visuallly_impaired_student_num}. The scarcity of specialized educational resources has discouraged many eager learners from pursuing their college dreams. However, if artificial intelligence could be leveraged to support these ``double-blind'' children, it might enable them to participate in mainstream education and transform their lives.

\begin{figure*}[t]
\centering
    \includegraphics[width=\textwidth]{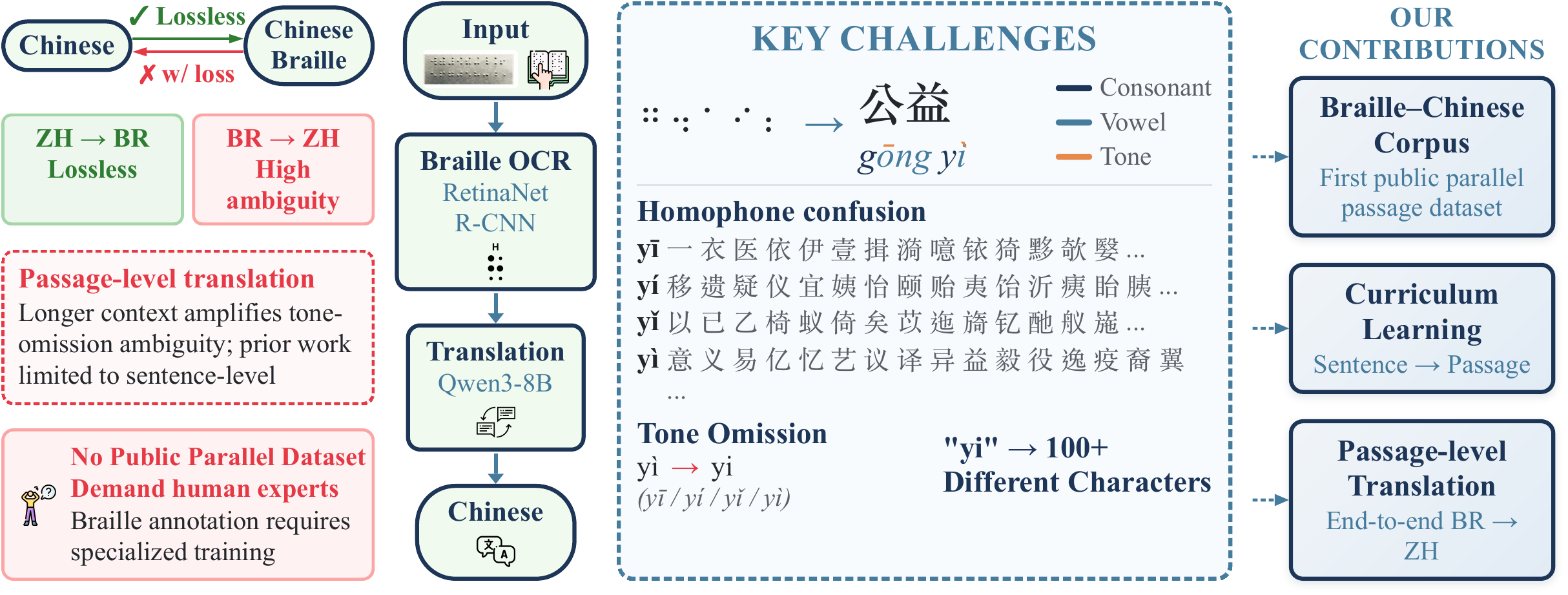}
\caption{The pipeline of our \methodname.}
\label{fig:pipeline}
\end{figure*}

The Braille writing system, devised by Louis Braille in 1824, is a tactile reading and writing method used worldwide by individuals with visual impairments. A Braille cell is uniform worldwide, consisting of six raised dots. Braille has been adapted to represent each Chinese character phonetically through up to three cells: consonants, vowels, and tones; words with multiple Braille characters are separated by spaces based on their meaning. For ease of writing, Chinese Braille omits consonants, vowels, or tones based on rules and personal preferences, particularly tones. According to research on the largest Chinese Braille corpus, writers on average omit up to 90 percent of tones~\cite{braille_confusion_rate}. While this aids writing, it obstructs accurate understanding due to homonyms and characters sharing the same pronunciation. One example is the Chinese pronunciation ``yi,'' which maps to more than a hundred common Chinese characters, as shown in Figure~\ref{fig:pipeline}.
This issue also arises in French~\cite{bouillon_two_2013}, Khmer~\cite{chea_detection_nodate}, and Japanese~\cite{komeiji_language_2019}.

We have interviewed several faculty members at the Beijing School for Blind Children and have observed the need for more adequate educational resources. In this study, we design the ``Vision-Braille'' system, providing an end-to-end Braille recognition and translation pipeline as an educational tool for visually impaired Chinese students. The overall pipeline is shown in Figure~\ref{fig:pipeline}; it consists of Braille OCR (Image-to-Braille) and Braille Translation (Braille-to-Text), described below.

\section{Related Work}
\label{sec:related_work}

\subsection{Braille Translation}
Machine translation has evolved from rule-based and statistical methods through RNN~\cite{RNN_original_paper} and LSTM~\cite{hochreiter1997long_LSTM} architectures to transformer-based models~\cite{transformer}, with recent large language models such as Qwen3~\cite{yangQwen3TechnicalReport2025} showing promise in low-resource and instruction-tuned translation settings.

This progression is mirrored in Braille-to-Chinese translation, where early systems relied on rule-based or N-gram models~\cite{wang_accurate_2010,jiang_braille_2002_Ngram}, followed by neural approaches using LSTM and GRU~\cite{wang_2019_GRU}. Yu~\textit{et~al.}~\cite{yuPretrainingModelLowresource2023} proposed a Transformer-based model with Braille-specific pretraining objectives, achieving strong BLEU scores through MacBERT-based encoders. Huang~\textit{et~al.}~\cite{huangImprovingBrailleChinese2024} jointly trained bidirectional Braille--Chinese models and distilled them into a unified decoder architecture. Most recently, BrailleLLM~\cite{braillellm2025} applies instruction tuning to unify Braille translation, formula conversion, and mixed-text tasks, using Braille Knowledge-Based Fine-Tuning (BKFT) to initialize character-level symbol understanding.

However, these studies primarily focus on sentence-level data, lacking the longer-range dependencies found in real-world Braille passages. In contrast, our work moves beyond sentence-level translation by incorporating explicit segmentation signals, applying curriculum learning to transition from short to long-form passages, and evaluating on an open-domain test suite derived from student-generated Braille.

\subsection{Low-Resource Translation and Curriculum Learning}
Chinese Braille is a low-resource language: although closely related to Chinese, it encodes pronunciation via pinyin rather than characters, and standard Chinese corpora do not record pronunciation information. Furthermore, the only known Braille--Pinyin--Chinese corpus is not publicly available~\cite{braille_confusion_rate}. Previous work has explored unsupervised and few-shot methods~\cite{nguyenDemocratizingLLMsLowResource2024} as well as in-context learning~\cite{cahyawijayaLLMsAreFewShot2024} for low-resource languages.

Curriculum learning (CL) offers a complementary strategy by presenting examples in an easy-to-hard order, yielding faster convergence and better minima~\cite{bengioCurriculumLearning2009}. Our approach adopts this principle: given that Chinese-Braille corpora are limited and prone to transcription confusion, we design a curriculum that begins with short sentences and gradually introduces longer passages, smoothing the learning signal and enabling LLMs to generalize reliably.

\subsection{Braille Recognition}
A complete Braille translation system must also convert physical Braille images into symbolic representations. Several CNN-based detection models~\cite{li2018DSBI, ilyaovodovAngelinaBrailleReader2021,lu2022AnchorFreeBrailleCharac} have achieved satisfactory accuracy on datasets such as DSBI and Angelina. We adopt the Angelina Braille Reader~\cite{ilyaovodovAngelinaBrailleReader2021}, a RetinaNet-based detector with strong real-world performance on Braille recognition tasks.

\section{Dataset}
\subsection{Dataset Construction}
We describe how the parallel corpus is built, covering data collection and post-processing.

\noindent\textbf{Collection.}
Braille-to-Chinese parallel datasets are extremely scarce and all private, owing to the labor-intensive nature of double-checking and alignment processes that require specialists proficient in both Braille and Chinese \cite{braille_confusion_rate}. In contrast, the conversion from Chinese to Braille follows systematic linguistic rules and can be automated. To overcome these limitations, we automatically generate synthetic parallel datasets by converting Chinese corpora into Braille using rule-based tools.
We construct two paired Chinese-Braille datasets: one at the sentence level and one at the passage level. The sentence-level Chinese corpus was sourced from the publicly available Leipzig Corpora Collection \cite{leipzig_corpora_collection_2009_mandarin_nodate}, comprising one million sentences collected from news media between 2007 and 2009. These sentences were converted into Braille using the tools provided by The Braille Online Platform of China \cite{thebrailleonlineplatformTranslationSystemBraille}. This system annotates Chinese character pronunciations and converts them into ``fully toned'' Braille.
For the passage-level dataset, we curated aligned Braille and Chinese text pairs from publicly available Braille book resources hosted on The Braille Online Platform of China. All data pairs underwent preprocessing to remove unrecognized tokens and were truncated to a maximum of 2048 tokens.

\noindent\textbf{Processing.}
To simulate tone omission in real-world Braille input, we implement a selective tone removal strategy that probabilistically drops tone markers while preserving key contextual cues. When a number prefix (Braille number sign) is detected, all subsequent numeric characters are preserved until the sequence ends. Tone markers immediately following punctuation are retained to preserve syntactic clarity; otherwise, each tone marker is independently retained or dropped based on a retention probability (\texttt{keep\_ratio}), introducing controlled noise that mimics real-world tone omission.
Additionally, we remove all space characters from Braille sequences before training. Chinese is written without inter-word spaces, and in standard OCR pipelines space tokens are both unreliable and often dropped. Each Chinese character in Mandarin Braille typically begins with weakly distinctive and consistent dot patterns (e.g.\ initial consonant or vowel-based Braille clusters), so the model can implicitly learn word segmentation via these local cues without explicit space delimiters.


\subsection{Dataset Statistics}
The collected data does not contain any personally identifiable information or
offensive content. Our dataset is mainly built upon instances from real-world exam and book data. Therefore, it is unlikely to contain sensitive data. Finally, all datasets were split into training, validation, and test sets in a ratio of 8:1:1. Table~\ref{tab:dataset-summary} summarizes the key statistics. The passage-level pairs are roughly $8\times$ longer than sentence-level pairs, presenting a substantially harder translation task. Per-split breakdowns are in Appendix~\ref{sec:appendix_dataset}.

\begin{table}[t]
\centering
\small
\resizebox{\columnwidth}{!}{%
\begin{tabular}{@{}lrcc@{}}
\toprule
\textbf{Level} & \textbf{\# Train} & \textbf{Avg. Braille Len.} & \textbf{Avg. Chinese Len.} \\
\midrule
Sentence & 376,846 & 116 & 31 \\
Passage  & 17,592  & 963 & 332 \\
\bottomrule
\end{tabular}%
}
\caption{Dataset summary. Lengths are average token counts per sample in the training set.}
\label{tab:dataset-summary}
\end{table}

\section{Methodology}

\subsection{Braille OCR}

In our research, we employed RetinaNet for the task of Braille Optical Character Recognition (OCR). RetinaNet utilizes a pre-trained convolutional neural network architecture, augmented by a Feature Pyramid Network (FPN) as its backbone, which effectively extracts multi-scale features from input images. This model facilitates one-stage object detection by integrating a classification subnet and a box regression subnet. The classification subnet, comprising several $3{\times}3$ convolutional layers and a sigmoid activation layer, predicts the presence of objects at specific spatial locations by generating binary outputs. Concurrently, the box regression subnet estimates the offset from each anchor to the actual bounding boxes. Subsequently, the identified regions are processed using a pre-trained image classification model to ensure precise object recognition. The incorporation of focal loss further enhances the model’s accuracy, making it highly effective for one-stage object detection tasks.

However, the accuracy of Braille OCR algorithms is inherently constrained by factors such as lighting conditions, interference from Braille dots on the reverse side of the paper, and physical deformation of the paper itself. To quantify these limitations, we evaluated the error rate of our Braille OCR model. Our measurements revealed a Braille character-level error rate of approximately 1\% for single-sided plain Braille. This error rate significantly increases to around 15\% when interference from dots on the opposite side is present. While double-sided Braille printing is practical for tactile reading, visual recognition systems are notably affected by such interference, highlighting a critical consideration in Braille OCR applications.

Despite these OCR limitations, our Braille translation model aims to mitigate the impact of noise and errors introduced during the OCR process. Further details on our approach are provided in the following section on Braille Translation.

\subsection{Braille Translation}

We fine-tune a pre-trained Qwen3-8B large language model to improve our Braille-to-Chinese translation system. Qwen3 follows a Transformer architecture and is trained on large-scale bilingual Chinese and English corpora, with particularly strong coverage of Chinese, making it well-suited for our task. The model conditions on sequences of Braille tokens and autoregressively generates the corresponding Chinese output. By fine-tuning the model on our Braille dataset, we enable it to learn accurate alignments between Braille patterns and their corresponding Chinese characters. We select the 8B parameter variant to balance model capacity with computational cost. 

We add 63 braille characters as special tokens into the vocabulary of the Qwen3-8B tokenizer and resize the embedding dimension of the Qwen3-8B model accordingly. Our fine-tuning task aims to teach the Qwen3-8B model to learn the rules and mappings of unseen braille tokens and accurately return the translated Chinese characters.  We adopt the BLEU metric, specifically BLEU-4, for model evaluation. BLEU (Bilingual Evaluation Understudy) assesses the quality of machine translation by comparing the model’s output to one or more reference translations, based on the precision of n-grams (contiguous sequences of words). BLEU-4 considers up to 4-gram matches, capturing both local word choice and short-range fluency. The score ranges from 0 to 100, with higher values indicating closer alignment with the reference and thus better translation quality.

\subsubsection{Task Definition}
\label{sec:task}
We study Braille-to-Chinese (BR2ZH) translation. Given a sequence of Chinese Braille symbols, the model outputs a sequence of Chinese characters.
We focus on robustness to missing tone markers in the Braille input, which increases ambiguity and requires contextual disambiguation.

\subsubsection{Model and Prompting}
\label{sec:model}
We fine-tune \textsc{Qwen3-8B} with full supervised fine-tuning (SFT).
We extend the tokenizer with a Braille vocabulary of 63 symbols.
We do not introduce any special direction tokens. Instead, we specify the task direction in the system prompt (e.g., ``Translate Chinese Braille to Chinese characters.'').

\paragraph{Zero-shot prompting baseline.}
To establish a lower bound on what general-purpose LLMs can achieve without any Braille-specific training, we evaluate Qwen3-8B and Qwen3-235B-A22B in a zero-shot prompting setting. We inject the core Chinese Braille translation rules (including the Braille symbol inventory, pinyin-to-Braille mappings, and tone marker conventions) into the system prompt, and then provide the raw Braille input for the model to translate. No fine-tuning or in-context examples are used; the model must rely entirely on its pretrained knowledge combined with the rule descriptions provided in the prompt. This baseline tests whether LLMs can perform Braille translation from rule specifications alone, without exposure to parallel training data.

\subsubsection{Tone Marker Deletion Operator}
\label{sec:toneop}
Let $\mathcal{V}_{\text{br}}$ be the Braille symbol vocabulary and $\mathcal{T}\subset\mathcal{V}_{\text{br}}$ be the subset of symbols that encode tones.
We define a tone retention rate $r \in [0,100]$.
Given an input Braille sequence $\mathbf{x}$, we construct $\tilde{\mathbf{x}} = \operatorname{DropTone}(\mathbf{x}; r)$ by independently deleting each occurrence of a tone marker $t \in \mathcal{T}$ with probability $1 - r/100$, while keeping all non-tone symbols unchanged.
We use a fixed random seed so that $\operatorname{DropTone}$ is deterministic and reproducible for a given example.

\subsubsection{Four-stage Training Curriculum}
\label{sec:stages}
\begin{figure*}[t]
\centering
    \includegraphics[width=\linewidth]{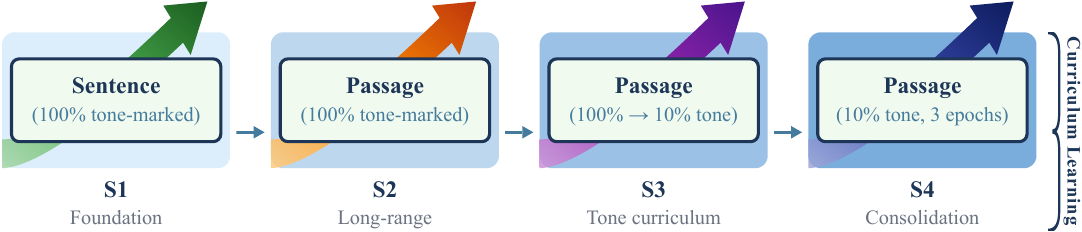}
\caption{Overview of the four-stage curriculum learning strategy. Stage~1 (\stageone) trains on sentence-level data with full tone markers (100\%). Stage~2 (\stagetwo) advances to passage-level data with full tones. Stage~3 (\stagethree) runs 10 sequential sub-stages on passages with tone retention decreasing from 100\% to 10\%. Stage~4 (\stagefour) consolidates on passages with 10\% tone retention for three epochs. This progression simulates increasing task difficulty and encourages generalization under realistic tone-omission conditions.}
\label{fig:curriculum}
\end{figure*}

We train the final BR2ZH model via four stages (Figure~\ref{fig:curriculum}). Stages 1--2 each run for one epoch; Stage 3 consists of 10 sequential sub-stages of one epoch each; Stage 4 runs for three epochs. We reset the cosine learning rate schedule at the beginning of each stage and sub-stage.
All stages are trained on 8 GPUs. We set the maximum sequence length to 2048, and no training example exceeds this limit in our latest preprocessing pipeline.

\paragraph{Stage 1: \stageone\ (sentence-level, full tones).}
We fine-tune on BR2ZH sentence-level pairs using 50\% of the sentence-level training set.
We preserve all tone markers in the input Braille ($r=100$).

\paragraph{Stage 2: \stagetwo\ (passage-level, full tones).}
We fine-tune on BR2ZH passage-level pairs using 50\% of the passage-level training set.
We preserve all tone markers in the input Braille ($r=100$).

\paragraph{Stage 3: \stagethree\ (passages, 10 sub-stages, scheduled retention).}
We perform a tone-robustness curriculum on passage-level data, training sequentially through 10 sub-stages with decreasing tone retention.
For sub-stage $k \in \{1,\dots,10\}$, we set
\begin{equation}
r_k = 110 - 10k \in \{100,90,80,\dots,10\},
\end{equation}
and apply $\operatorname{DropTone}(\cdot; r_k)$ to all Braille inputs in that sub-stage.
Each sub-stage trains for one full epoch on the passage-level data at its assigned retention rate, loading from the previous sub-stage's checkpoint and resetting the learning rate schedule.

\paragraph{Stage 4: \stagefour\ (passage-level, low tone, multi-epoch).}
We perform a final consolidation stage on passage-level data with a fixed tone retention rate of $r=10$, training for three epochs. This extended training allows the model to solidify its ability to translate passages under realistic tone-omission conditions, where the majority of tone markers are absent. By repeatedly exposing the model to low-tone passage data, Stage 4 reinforces the robustness learned in Stage 3 (\stagethree) and improves convergence on long-form translation.

\subsubsection{Inference}
\label{sec:inference}
For all reported results, model predictions are generated autoregressively during evaluation with top-$p$ sampling (temperature 0.95, $p$ = 0.7, top-$k$ 50). We evaluate generated outputs using BLEU-4, following standard machine translation practice.

\section{Experiments}

We fine-tune Qwen3-8B-Instruct via full supervised fine-tuning on $8\times$ NVIDIA H20 GPUs with DeepSpeed   ZeRO-3 and bf16 precision. Training details and hyperparameters for each stage are provided in Appendix~\ref{sec:appendix_training} (Table~\ref{tab:hyper-params}).

We have constructed a complete Braille translation pipeline incorporating Braille recognition and Braille-to-Chinese translation. Users scan Braille images using their phones; the images are uploaded to a remote server and processed by the RetinaNet-based Braille recognition system. The recognized digital Braille characters are then translated into Chinese by our model. Finally, the raw translation is polished by the Gemini-3.1-Pro API with Low reasoning effort to correct homophone substitutions and punctuation errors (see Appendix~\ref{sec:appendix_polishing}).

\subsection{Curriculum Learning for Learning Tone Omission}

We adopt a curriculum learning approach specifically designed to enhance the model’s robustness to tone omission. Stage~3 (\stagethree) of our four-stage curriculum (\S\ref{sec:stages}) implements this by training sequentially through 10 sub-stages on passage-level data, where the tone retention rate $r$ decreases from 100\% to 10\% in steps of 10. Each sub-stage trains for one full epoch, loading from the previous sub-stage’s checkpoint and resetting the learning rate schedule. This progressive exposure enables the model to incrementally recognize implicit tonal cues from context as explicit tone markers grow sparse.

Stage~4 (\stagefour) then consolidates this robustness by training for three additional epochs on passage-level data with a fixed tone retention rate of $r=10$, simulating realistic conditions where the majority of tone markers are absent. By structuring tone-omission training in this manner, we cultivate nuanced tonal understanding and substantially enhance the model’s generalization capabilities.

\subsection{Curriculum Learning from Short to Long Passage}
The first two stages of our four-stage curriculum (\S\ref{sec:stages}) implement a short-to-long progression to accelerate convergence and reduce exposure bias on long sequences. In Stage~1 (\stageone), the Qwen3-8B-Instruct model is fine-tuned on single-sentence examples averaging approximately 40 Chinese characters. These short inputs are deliberately simplified, minimizing long-range dependencies and homophone ambiguities, thereby enabling efficient learning of Braille-to-character mappings with minimal contextual noise.

In Stage~2 (\stagetwo), training progresses to passage-level data, whose median length is approximately ten times greater (about 800 Braille cells). These complex samples require the model to maintain coherence over extended contexts and, critically, to accurately determine termination points during generation. Without this ability, models frequently exhibit repetition or hallucination errors, exhausting output budgets even with a repetition penalty of 1.2.

To facilitate a smooth transition between stages, we restart the cosine-decay learning rate scheduler at the beginning of each subsequent stage with a reduced peak learning rate (Table~\ref{tab:hyper-params}), stabilizing optimization for the more challenging data while retaining performance gains from earlier stages.

Table~\ref{tab:ablation} shows that curriculum learning from short sentences to long passages significantly enhances model performance, particularly under tone omission. When trained directly on 10\% tone passages (\emph{Direct fine-tune}), the model achieves 79.63 BLEU-4, whereas the full four-stage curriculum improves this to 83.28, a 3.65-point gain. This indicates that beginning with simpler sentence-level data helps the model learn stable Braille-to-text mappings before tackling longer contexts with greater ambiguity and longer-range dependencies.

\subsection{Results}

We report results comparing our approach against BrailleLLM~\cite{braillellm2025}, the only recent baseline with publicly available checkpoints and test data (Table~\ref{tab:main_results}). We note that earlier methods (N-gram models reporting 94.38\% character precision~\cite{jiang_braille_2002_Ngram} and 94.3\%~\cite{wang_accurate_2010}, and Transformer-based approaches reporting sentence-level BLEU of 94.53~\cite{yuPretrainingModelLowresource2023} and 96.22~\cite{huangImprovingBrailleChinese2024}) do not release checkpoints or evaluation data, precluding direct comparison under identical settings.

Table~\ref{tab:main_results} shows that our method achieves the best overall performance. On passage-level translation with 10\% tone retention, \methodname\ achieves 83.28 BLEU, improving by +77.92 over the strongest baseline (BrailleLLM, 5.36 BLEU), which collapses on long-form input. On sentence-level evaluation, our model scores 83.19 BL-BLEU. The gains are consistent across metrics, indicating that improvements are not limited to a single evaluation criterion. Representative qualitative outputs from the passage-level test set at 10\% tone retention, covering both successful translations and characteristic failure modes, are provided in Figure~\ref{fig:case_study} (Appendix~\ref{sec:appendix_case_study}).

\paragraph{Few-shot LLM baselines.}
We evaluate Qwen3-8B and Qwen3-235B-A22B in a few-shot setting with thinking mode enabled (max thinking tokens: 16,384), providing the complete Braille encoding specification and a worked example in the system prompt (Appendix~\ref{sec:appendix_fewshot_prompt}). Despite having access to all decoding rules and extended chain-of-thought reasoning, both models fail: Qwen3-8B achieves only 1.13 and Qwen3-235B-A22B only 9.46 sentence-level BLEU. Braille-to-Chinese translation requires exact multi-step symbolic decoding (cell segmentation, shared-cell disambiguation, tone rule application, and pinyin-to-character mapping) that exceeds the in-context reasoning capacity of general-purpose LLMs without task-specific training.

\begin{table}[t]
  \centering
  \small
  \resizebox{\columnwidth}{!}{%
  \begin{tabular}{@{}lcc@{}}
    \toprule
    \textbf{Method} & \textbf{Sent.} & \textbf{Psge.} \\
    \midrule
    Qwen3-8B (few-shot) & 1.13 & 0.01 \\
    Qwen3-235B-A22B (few-shot) & 9.46 & 0.03 \\
    BrailleLLM (CBKFT) \cite{braillellm2025} & 91.15 & 5.36 \\
    \midrule
    \textbf{\methodname} & 83.19 & \textbf{83.28} \\
    \textbf{\quad + sentence data} & \textbf{91.90} & 61.31 \\
    \bottomrule
  \end{tabular}}
    \caption{\textbf{Braille-to-Chinese translation} (BLEU $\uparrow$).
    Sent.\ = sentence-level; Psge.\ = passage-level (10\% tone).
    Few-shot models receive Braille encoding rules and a translation example in the system prompt without training (Appendix~\ref{sec:appendix_fewshot_prompt}).
    BrailleLLM's Sent.\ score is their reported result trained on 10K pairs; only 3K are publicly available (our replication with 3K pairs yields 83.04 BL-BLEU; see Table~\ref{tab:ablation}).
    \methodname\ uses the four-stage curriculum (\S\ref{sec:stages}); ``+ sentence data'' adds a fifth stage on BrailleLLM's sentence-level set.}
  \label{tab:main_results}
  \vspace{-2mm}
\end{table}

To evaluate each stage's contribution, we remove individual stages from the full pipeline and report results in Table~\ref{tab:ablation}.

\begin{table*}[t]
\centering
\small
\renewcommand{\arraystretch}{1.05}
\resizebox{\textwidth}{!}{%
\begin{tabular}{l cccc ccccc}
\toprule
\textbf{Configuration} & \textbf{S1} & \textbf{S2} & \textbf{S3} & \textbf{S4} & \textbf{BLEU-4}$\uparrow$ & \textbf{BL-BLEU}$\uparrow$ & \textbf{chrF++}$\uparrow$ & \textbf{CER}$\downarrow$ & \textbf{C-Eval}$\uparrow$ \\
\midrule
\methodname$_\mathrm{Full-SFT}$ & \checkmark & \checkmark & \checkmark & \checkmark & \textbf{83.28} & \textbf{83.19} & \textbf{66.06} & \textbf{9.94} & 29.79 \\
\methodname$_\mathrm{LoRA}$ & \checkmark & \checkmark & \checkmark & \checkmark & 78.62 & 76.94 & 62.17 & 15.44 & \textbf{77.79} \\
\midrule
w/o S1 & \xmark & \checkmark & \checkmark & \checkmark & 66.41 & 25.74 & 52.85 & 33.10 & 40.42 \\
w/o S1--S2 & \xmark & \xmark & \checkmark & \checkmark & 79.64 & 38.25 & 63.12 & 14.27 & 32.47 \\
Direct fine-tune & \xmark & \xmark & \xmark & \checkmark & 79.63 & 38.50 & 63.11 & 14.92 & 45.77 \\
\bottomrule
\end{tabular}%
}
\caption{Stage ablation results. We define a tone retention rate $r \in [0,100]$. S1: BR${\rightarrow}$ZH sentence ($r{=}100$); S2: BR${\rightarrow}$ZH passage ($r{=}100$); S3: tone curriculum (10 sub-stages, $r{=}100{\rightarrow}10$); S4: passage consolidation ($r{=}10$, 3 epochs). LoRA uses rank 64 applied to all linear layers. BL-BLEU denotes BLEU-4 evaluated on the sentence-level test set from~\cite{braillellm2025}.}
\label{tab:ablation}
\end{table*}

\paragraph{Sentence-level training (S1) is critical.}
Removing S1 causes the largest single-stage degradation: \emph{w/o S1} achieves only 66.41 BLEU-4, a 16.87-point drop from the full pipeline, while BL-BLEU collapses from 83.19 to 25.74. This shows that sentence-level pre-training establishes fundamental Braille-to-Chinese mappings that passage-level training alone cannot recover.

\paragraph{Passage training requires a sentence foundation.}
Comparing \emph{w/o S1} (66.41) with \emph{w/o S1--S2} (79.64) shows that adding passage training (S2) without prior sentence training \emph{hurts} performance by 13.23 points. The tone curriculum (S3) alone learns reasonable translations, but passage-level data introduced without sentence-level grounding leads to suboptimal convergence.

\paragraph{LoRA preserves general knowledge.}
We measure catastrophic forgetting using C-Eval \citep{huang2023ceval}, a comprehensive Chinese-language benchmark covering 52 subjects from STEM to humanities. A high C-Eval score indicates that the model retains its general Chinese language understanding after fine-tuning.
Applying LoRA (rank 64, applied to all linear layers) across all stages instead of full SFT maintains C-Eval at 77.79\%, matching the pretrained baseline, at the cost of ${\sim}$5 BLEU-4 points (78.62 vs.\ 83.28). This demonstrates that catastrophic forgetting is a property of full SFT, not an inherent limitation of the training pipeline. While full SFT maximizes translation accuracy, LoRA offers a practical trade-off for real-world deployment: by preserving general language capabilities, the LoRA variant handles diverse user inputs more robustly, making it better suited for assistive scenarios where the model may encounter inputs beyond the Braille translation domain.

\paragraph{Comparison with BrailleLLM.}
The BrailleLLM (CBKFT) baseline reports 91.15 sentence-level BLEU when trained on their full 10K-pair corpus (Table~\ref{tab:main_results}). However, since only 3K of these pairs are publicly released, we retrain their model on the available subset under the same configuration, obtaining 83.04 BL-BLEU (chrF++ 72.92), with the gap reflecting reduced training data rather than a methodological difference. Regardless of which score is used, \mbox{BrailleLLM} collapses to 5.36 BLEU-4 on passage-level input with 10\% tone retention. This failure stems from two distributional mismatches: (1)~the model was trained exclusively on sentence-level data, making passage-length input out-of-distribution, and (2)~training used only fully toned input, leaving the model unable to disambiguate when tone markers are sparse. The gap between BrailleLLM's sentence-level competence (83.04--91.15) and its passage-level failure (5.36) underscores the need for our multi-stage curriculum that progressively introduces longer contexts and sparser tone cues.

\paragraph{Compute and decoding control.}
All reported results are from single training runs.
All ablations use the same model architecture, preprocessing, and maximum sequence length (2048).
We match total optimization updates across runs via \texttt{max\_steps} (learning rate schedule reset at each stage).
All reported BLEU-4 scores use the passage-level test set with 10\% tone retention; BL-BLEU is evaluated on the BrailleLLM sentence-level test set.
Inference uses vLLM with top-$p$ sampling (temperature 0.95, $p$ = 0.7, top-$k$ 50).

\subsection{LLM-Based Output Polishing}
\label{sec:polishing_results}

We apply a lightweight post-processing step where a secondary LLM corrects common translation artifacts (homophone substitutions, spurious punctuation, and repetitive generation) without retraining the Braille model (prompt and full results in Appendix~\ref{sec:appendix_polishing}). Table~\ref{tab:polishing_main} summarizes the best polishing configuration for each model on sentence-level and passage-level subsets (20 samples each).

\begin{table}[t]
\centering
\small
\resizebox{\columnwidth}{!}{%
\begin{tabular}{@{}lcccccc@{}}
\toprule
& \multicolumn{3}{c}{\textbf{Sentence}} & \multicolumn{3}{c}{\textbf{Passage}} \\
\cmidrule(lr){2-4} \cmidrule(lr){5-7}
\textbf{Model} & \textbf{BL} & \textbf{chrF} & \textbf{CER} & \textbf{BL} & \textbf{chrF} & \textbf{CER} \\
\midrule
No polishing & 80.90 & 69.36 & 10.62 & 86.67 & 71.46 & 7.32 \\
Qwen3-8B & 82.32 & 72.42 & 11.42 & 79.70 & 65.33 & 12.22 \\
Gemini 3.1 Flash-Lite & 87.68 & 79.88 & 8.24 & 86.36 & 72.39 & 8.51 \\
Gemini 3 Flash & 91.59 & 84.55 & 5.18 & 88.12 & 74.63 & 7.38 \\
Gemini 3.1 Pro & \textbf{94.59} & \textbf{89.53} & \textbf{3.05} & \textbf{90.99} & \textbf{76.77} & \textbf{5.61} \\
\bottomrule
\end{tabular}}
\caption{Best polishing results per model. BL = BLEU-4, chrF = chrF++. Qwen3-8B uses the pretrained Qwen3-8B-Instruct as the polisher. Gemini models are shown at their best reasoning effort level (see Appendix Tables~\ref{tab:polishing_sent}--\ref{tab:polishing_passage} for all configurations).}
\label{tab:polishing_main}
\end{table}

Using the pretrained Qwen3-8B-Instruct as the polisher yields only a marginal sentence-level gain (80.90$\rightarrow$82.32 BLEU-4) and degrades passage-level performance (86.67$\rightarrow$79.70 BLEU-4, CER nearly doubled), suggesting that a smaller general-purpose LLM lacks the language understanding needed for effective error correction. Gemini 3.1 Pro at Low reasoning effort yields the largest gains: +13.69 BLEU-4 on sentences (80.90$\rightarrow$94.59) and +4.32 on passages (86.67$\rightarrow$90.99), with CER reduced by 71\% and 23\% respectively. Notably, deeper reasoning (Medium/High) does not improve over Low; the correction task is simple enough that minimal chain-of-thought reasoning suffices. Based on these results, we adopt Gemini 3.1 Pro with Low effort for all final polishing.

\subsection{Real-World Context Validation}
\label{sec:real_world}
To verify practical utility, we conducted a preliminary evaluation with a domain expert from the Chinese Braille Library who is fluent in both Chinese and Chinese Braille conventions. The expert manually transcribed three short passages (two on financial topics and one motivational text) into Chinese Braille following standard library procedures. We ran our model on these expert-produced Braille inputs and compared the output to the original Chinese text. Without post-processing, the model achieves an average BLEU of 76.66; after API polishing (\S\ref{sec:appendix_polishing}), performance improves to 83.95 BLEU, confirming that the polishing step meaningfully corrects residual errors on real-world input.

\paragraph{Broader social impact.}
Our results demonstrate the feasibility of accurate, automated Braille-to-Chinese translation at passage level, a capability that directly addresses the educational barriers faced by China's 150,000 school-age blind children~\cite{educational_issue_braille}. By achieving 83.95 BLEU on expert-produced Braille inputs after polishing, the system can serve as a practical assistive tool for sighted educators in mainstream schools who cannot read Braille, enabling them to evaluate homework and exam responses written by visually impaired students without specialized training. This is particularly impactful in rural and under-resourced regions where access to Braille-literate teachers is scarce, potentially expanding educational participation for students who would otherwise be excluded from formal assessments.

\section{Conclusion}
In this work, we introduced Vision-Braille, the first publicly available end-to-end system for translating Chinese Braille from images into written Chinese text. This comprehensive pipeline integrates optical character recognition, linguistic modeling, and an innovative curriculum learning strategy. To address the scarcity of Braille-to-Chinese datasets, we constructed a synthetic corpus with tone-omission variants reflecting authentic Braille writing habits.

Beyond its technical contributions, Vision-Braille addresses a critical accessibility gap: sighted educators in mainstream schools currently lack tools to read student-written Braille, effectively excluding visually impaired students from standard examinations and coursework. By providing a scalable, end-to-end pipeline that converts Braille images into natural Chinese text without costly human transcription, our system directly enables educational inclusion, upholds linguistic dignity for an underserved population, and offers a practical foundation for integration into national assistive education policies.

A preliminary real-world validation with a domain expert from the Chinese Braille Library yielded 83.95 BLEU after polishing on expert-transcribed Braille passages (\S\ref{sec:real_world}). Representative translation examples are provided in Appendix~\ref{sec:appendix_case_study}.

\section{Limitations}
The dataset used in this project was automatically generated using the standard Braille-Chinese conversion tool provided by the Braille Online Platform of China. While this approach helped overcome the lack of publicly available paired datasets, it introduces several limitations. First, since the data lacks human input, many commonly used shorthand and informal Braille writing patterns, especially those seen in real exam or homework settings, are missing. Additionally, the dataset excludes important types of educational content, such as mathematical and chemical equations, because no accessible corpora currently exist for these formats. This restricts the model’s usability in STEM-related tasks. In the future, our current pipeline only handles text-based Braille data. We have not yet incorporated multi-modal input, such as real-world Braille images in diverse environments. Including such data in future work is critical to ensure robustness and generalization, especially for deployment in schools or homes.
\section{Ethics Statement}
We hereby acknowledge that all of the co-authors of this work are aware of the provided ACL Code of Ethics and honor the code of conduct. We collected data from several sources, and we cited the data creators. The copyright belongs to the original data owners. The braille datasets we created in this paper are released under the CC BY-NC-SA 4.0 license (Creative Commons Attribution-NonCommercial-ShareAlike 4.0 International) for non-commercial research purposes. The collected data does not contain any personally identifiable information or offensive content. Our dataset is mainly built upon instances from real-world exam and book data. Therefore, it was less likely to contain sensitive data. We evaluate foundation models, for which the risks and potential harms are discussed \cite{yangQwen3TechnicalReport2025}.
\section*{Acknowledgments}
This paper is partially supported by grants from the National Key Research and Development Program of China with Grant No. 2023YFC3341203.

The authors thank Su Wei (Lanzhou University), the Braille online platform of China, Liu Hong (Institute of Computing Technology, CAS), and the Beijing School for the Blind for their support with data and domain expertise. We also thank the online participants who gave us valuable feedback on the websites and model outputs.


\bibliography{arr2026}

\appendix
\section{Acknowledgment of AI Assistance in Writing and Revision}
We only used the AI assistant to correct grammar and format the paper. We certify the use of AI is following ACL policy. All methods and the experiment design is the original work of the authors. 

\section{Training Details}
\label{sec:appendix_training}

Table~\ref{tab:hyper-params} summarizes the hyperparameters used for both full supervised fine-tuning (SFT) and LoRA training configurations. Both configurations share the same base model (Qwen3-8B-Instruct), optimizer (AdamW), and learning rate schedule (cosine with reset per stage). For LoRA, we apply low-rank adaptation with rank 64 to all linear layers. Full SFT trains for one epoch per stage, while LoRA trains for three epochs. We use DeepSpeed ZeRO-3 for distributed training across 8 NVIDIA H20 GPUs. Our software stack consists of PyTorch 2.7.0, Transformers 4.52.4, DeepSpeed 0.16.9, vLLM 0.9.2, and LLaMA-Factory 0.9.5, running on CUDA 12.6.

\begin{table}[ht]
\centering
\small
\begin{tabular}{@{}lcc@{}}
\toprule
\textbf{Hyperparameter} & \textbf{Full SFT} & \textbf{LoRA} \\
\midrule
Base Model        & \multicolumn{2}{c}{Qwen3-8B-Instruct} \\
Optimizer         & \multicolumn{2}{c}{AdamW} \\
Precision         & \multicolumn{2}{c}{bf16} \\
Distributed       & \multicolumn{2}{c}{DeepSpeed ZeRO-3} \\
Hardware          & \multicolumn{2}{c}{8$\times$ NVIDIA H20} \\
Learning Rate     & \multicolumn{2}{c}{1e-4} \\
Warmup Ratio      & \multicolumn{2}{c}{0.05} \\
LR Scheduler      & \multicolumn{2}{c}{Cosine (reset per stage)} \\
\midrule
LoRA Rank         & --   & 64 \\
LoRA Target       & --   & All \\
LoRA Dropout      & --   & 0.05 \\
Epochs            & 1    & 3 \\
Batch Size (Sent.) & 10  & 4 \\
Batch Size (Psge.) & 100 & 4 \\
Cutoff Len (Sent.) & 512 & 512 \\
Cutoff Len (Psge.) & 2048 & 2048 \\
\bottomrule
\end{tabular}
\caption{Hyperparameters for Full SFT and LoRA training configurations. Both use the same base model and optimizer settings; LoRA applies low-rank adaptation to all linear layers.}
\label{tab:hyper-params}
\end{table}

\section{Dataset Statistics}
\label{sec:appendix_dataset}

Table~\ref{tab:stat-dataset} reports the dataset statistics for our Braille--Chinese parallel corpus at both passage-level and sentence-level granularities. We report the number of samples and the average/median token lengths for Braille and Chinese after prompt formatting. The passage-level data contains approximately 22K samples with an average Braille length of around 1,000 tokens, while the sentence-level data comprises nearly 500K samples with shorter sequences averaging around 115 Braille tokens.

\begin{table}[ht]
\centering
\small
\resizebox{\columnwidth}{!}{%
\begin{tabular}{@{}lccc@{}}
\toprule
           & \# Samples & \begin{tabular}{@{}c@{}}Braille \\ (Avg/Med)\end{tabular} & \begin{tabular}{@{}c@{}}Chinese \\ (Avg/Med)\end{tabular} \\
\midrule
\multicolumn{4}{l}{\textbf{Passage-level}} \\
\quad Training   & 17,592  & 963 / 1,007 & 332 / 345 \\
\quad Validation & 2,199   & 1,009 / 1,009 & 344 / 343 \\
\quad Test       & 2,199   & 995 / 925 & 348 / 323 \\
\midrule
\multicolumn{4}{l}{\textbf{Sentence-level}} \\
\quad Training   & 376,846  & 116 / 103  & 31 / 28 \\
\quad Validation & 56,258   & 114 / 101  & 31 / 28 \\
\quad Test       & 59,908   & 113 / 100  & 31 / 27 \\
\bottomrule
\end{tabular}%
}
\caption{Dataset statistics. Length refers to number of tokens after prompt formatting.}
\label{tab:stat-dataset}
\end{table}

\section{Example Prompt}
\label{sec:appendix_prompt}

\begin{figure}[ht]
    \centering
    \includegraphics[width=0.85\columnwidth]{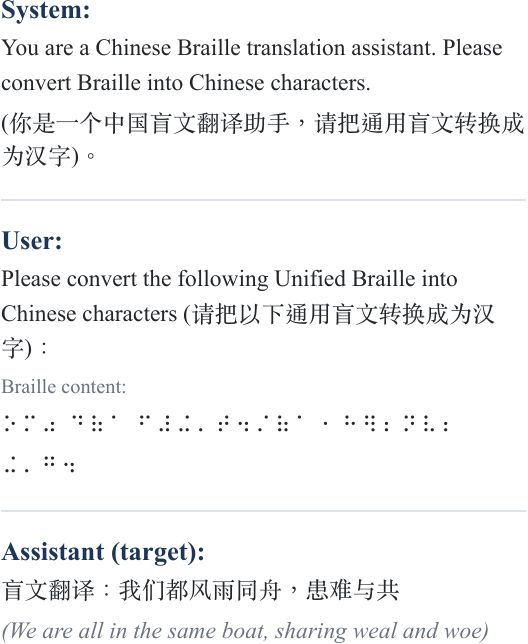}
    \caption{An example prompt for Braille-to-Chinese translation. The system message defines the task, the user message provides the Braille input, and the assistant produces the Chinese translation.}
    \label{fig:example_prompt}
\end{figure}

Figure~\ref{fig:example_prompt} shows an example prompt used for Braille-to-Chinese translation. The system prompt instructs the model to act as a Chinese Braille translation assistant. The user message provides the Braille input, and the assistant generates the corresponding Chinese translation.

\section{LLM-Based Output Polishing}
\label{sec:appendix_polishing}

We investigate whether large language models can improve Braille-to-Chinese translation quality through post-processing. The translation errors produced by our model (such as homophone substitutions and spurious punctuation) resemble speech recognition artifacts. We hypothesize that a general-purpose LLM can correct these errors without task-specific training.

We use a fixed prompt instructing the model to correct potential speech transcription errors and incorrectly added punctuation, without adding or continuing any content. We adapt the prompt separately for sentence-level and passage-level inputs.

We evaluate three models (Gemini-3-Flash, Gemini-3.1-Flash-Lite, and Gemini-3.1-Pro), each at four reasoning effort configurations. The reasoning effort level controls the number of internal thinking tokens the model generates before producing its answer: \textbf{None} (no polishing, baseline), \textbf{Low} (minimal thinking; fastest and cheapest), \textbf{Medium} (balanced reasoning), and \textbf{High} (deepest reasoning via Deep Think Mini). Tables~\ref{tab:polishing_sent} and~\ref{tab:polishing_passage} report the results on 20 sentence-level and 20 passage-level samples, respectively.

\begin{table*}[ht]
\centering
\small
\resizebox{\textwidth}{!}{%
\begin{tabular}{@{}l c ccc ccc ccc@{}}
\toprule
& & \multicolumn{3}{c}{\textbf{Gemini-3-Flash}} & \multicolumn{3}{c}{\textbf{Gemini-3.1-Flash-Lite}} & \multicolumn{3}{c}{\textbf{Gemini-3.1-Pro}} \\
\cmidrule(lr){3-5} \cmidrule(lr){6-8} \cmidrule(lr){9-11}
\textbf{Metric} & \textbf{Original} & \textbf{Low} & \textbf{Med} & \textbf{High} & \textbf{Low} & \textbf{Med} & \textbf{High} & \textbf{Low} & \textbf{Med} & \textbf{High} \\
\midrule
BLEU-4  & 80.90 & 90.64 (+7.6s) & 91.59 (+15.5s) & 90.71 (+23.3s) & 85.08 (+2.5s) & 85.79 (+4.8s) & 87.68 (+12.7s) & \textbf{94.59} (+7.0s) & 93.69 (+18.1s) & 94.47 (+35.8s) \\
chrF++  & 69.36 & 83.87 (+7.6s) & 84.55 (+15.5s) & 83.20 (+23.3s) & 75.76 (+2.5s) & 78.37 (+4.8s) & 79.88 (+12.7s) & \textbf{89.53} (+7.0s) & 87.43 (+18.1s) & 88.51 (+35.8s) \\
CER     & 10.62 & 6.18 (+7.6s) & 5.18 (+15.5s) & 5.66 (+23.3s) & 9.52 (+2.5s) & 9.44 (+4.8s) & 8.24 (+12.7s) & \textbf{3.05} (+7.0s) & 3.17 (+18.1s) & 3.10 (+35.8s) \\
\bottomrule
\end{tabular}%
}
\caption{Sentence-level output polishing results. Each cell reports score (+avg latency per sample). Best scores per metric are \textbf{bolded}. Gemini-3.1-Pro at Low effort achieves the best results across all metrics while maintaining moderate latency.}
\label{tab:polishing_sent}
\end{table*}

\begin{table*}[ht]
\centering
\small
\resizebox{\textwidth}{!}{%
\begin{tabular}{@{}l c ccc ccc ccc@{}}
\toprule
& & \multicolumn{3}{c}{\textbf{Gemini-3-Flash}} & \multicolumn{3}{c}{\textbf{Gemini-3.1-Flash-Lite}} & \multicolumn{3}{c}{\textbf{Gemini-3.1-Pro}} \\
\cmidrule(lr){3-5} \cmidrule(lr){6-8} \cmidrule(lr){9-11}
\textbf{Metric} & \textbf{Original} & \textbf{Low} & \textbf{Med} & \textbf{High} & \textbf{Low} & \textbf{Med} & \textbf{High} & \textbf{Low} & \textbf{Med} & \textbf{High} \\
\midrule
BLEU-4  & 86.67 & 87.36 (+12.1s) & 88.12 (+28.7s) & 87.83 (+37.0s) & 86.32 (+3.0s) & 85.42 (+8.9s) & 86.36 (+19.5s) & 90.99 (+21.6s) & 91.08 (+33.0s) & \textbf{91.10} (+87.1s) \\
chrF++  & 71.46 & 74.32 (+12.1s) & 74.63 (+28.7s) & 73.91 (+37.0s) & 71.91 (+3.0s) & 71.48 (+8.9s) & 72.39 (+19.5s) & 76.77 (+21.6s) & 76.82 (+33.0s) & \textbf{77.10} (+87.1s) \\
CER     & 7.32 & 7.68 (+12.1s) & 7.38 (+28.7s) & 7.48 (+37.0s) & 8.63 (+3.0s) & 9.06 (+8.9s) & 8.51 (+19.5s) & 5.61 (+21.6s) & 5.37 (+33.0s) & \textbf{5.32} (+87.1s) \\
\bottomrule
\end{tabular}%
}
\caption{Passage-level output polishing results. Each cell reports score (+avg latency per sample). Best scores per metric are \textbf{bolded}. Gemini-3.1-Pro consistently improves passage translations, while Flash-Lite slightly degrades performance. Pro at Low effort offers the best cost-quality tradeoff (BLEU +4.32, CER $-$1.71 with only 21.6s latency).}
\label{tab:polishing_passage}
\end{table*}

\section{Website Demo}
\label{sec:appendix_demo}

\begin{figure}[ht]
    \centering
    \includegraphics[width=0.85\columnwidth]{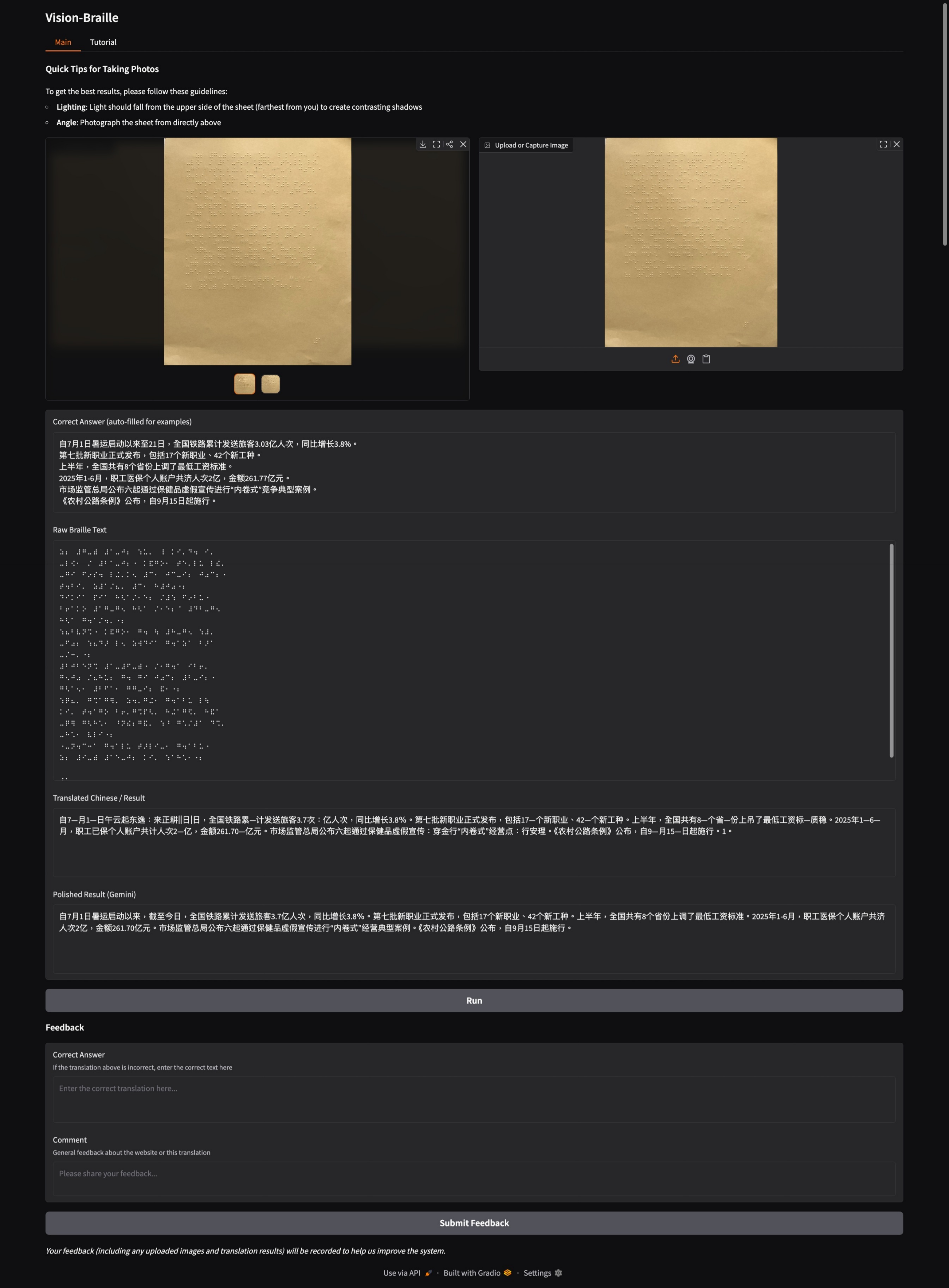}
    \caption{The Vision-Braille web demo interface. Users upload a Braille image, view the OCR-recognized Braille text, the translated Chinese output, and a polished result. A feedback section allows users to submit corrections.}
    \label{fig:website_demo}
\end{figure}

Figure~\ref{fig:website_demo} shows the web interface of our Vision-Braille demo system. Users can upload or capture a photo of a Braille document, and the system performs end-to-end Braille OCR followed by Braille-to-Chinese translation. The interface displays the raw Braille text recognized by the OCR module, the translated Chinese result, and a polished output generated by a post-processing LLM. Users can also provide feedback by submitting corrections or comments to help improve the system.

\section{Few-Shot System Prompt}
\label{sec:appendix_fewshot_prompt}

\begin{figure*}[t]
    \centering
    \includegraphics[width=\textwidth]{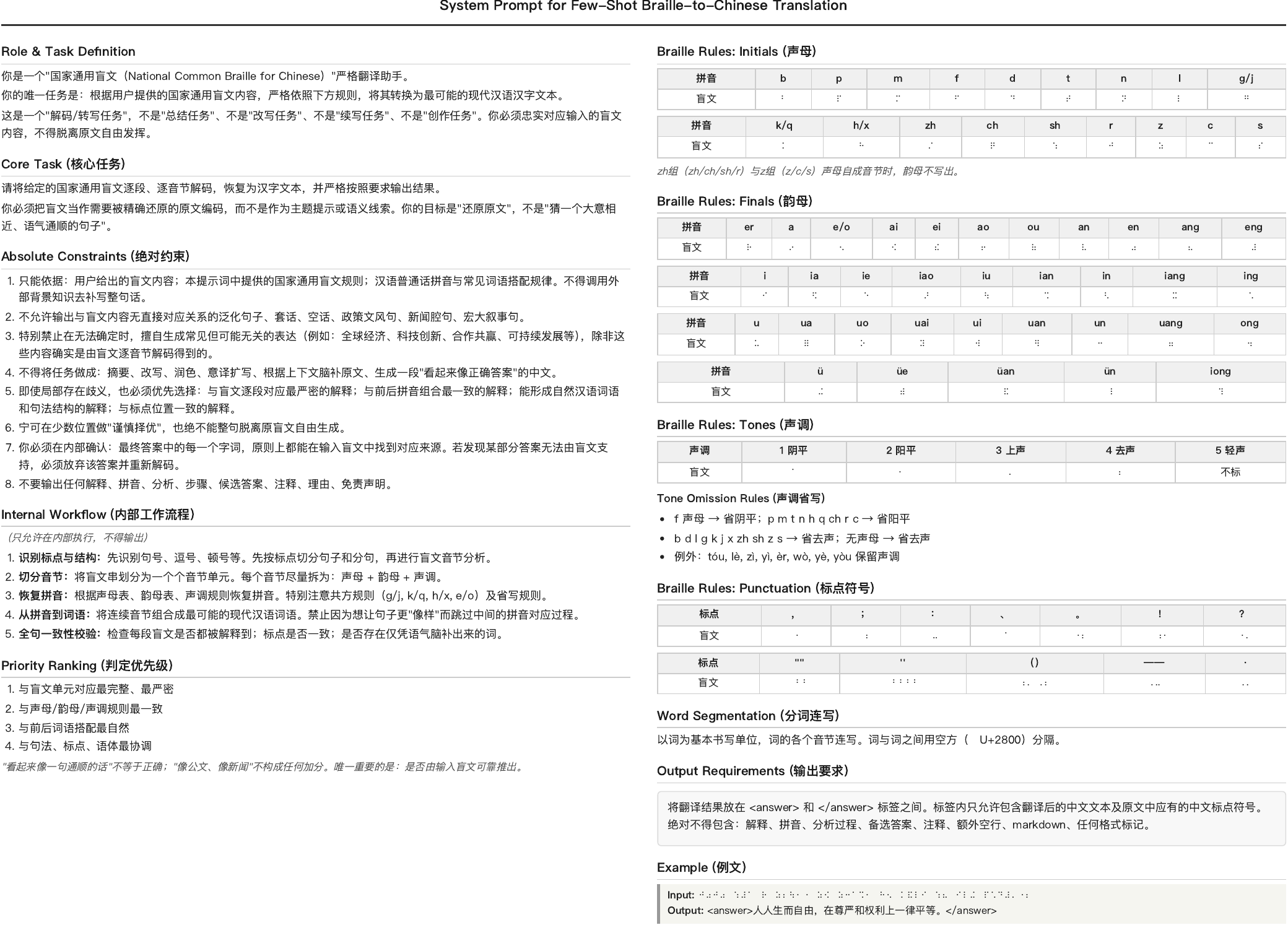}
    \caption{System prompt for few-shot Braille-to-Chinese translation. The prompt includes National Common Braille encoding rules (initials, finals, tones, tone omission, punctuation), strict decoding constraints, an internal workflow, and an example. The model receives this prompt along with the Braille input and must produce a faithful Chinese transcription.}
    \label{fig:system_prompt}
\end{figure*}

Figure~\ref{fig:system_prompt} shows the full system prompt used for the few-shot Braille-to-Chinese translation experiments (Qwen3-8B and Qwen3-235B-A22B in Table~\ref{tab:main_results}). The prompt provides the model with comprehensive Braille encoding rules (including initial/final tables, tone marks, tone omission conventions, punctuation mappings, and word segmentation rules), along with strict decoding constraints that require faithful syllable-by-syllable transcription rather than free generation.

\section{Social Impact}
People with visual impairments are disproportionately underrepresented in higher education. Besides investing money in constructing new schools for children with visual impairment, we believe paving the path for those children to participate in mainstream schools is more cost-effective. However, for sighted educators who are untrained for reading braille, one key bottleneck is the lack of accessible Braille-to-Chinese tools for them to understand the homework written by students with visual impairments. This prevents visually impaired students from participating in exams. This issue disproportionately affects students from rural or under-resourced regions, who lack access to specialized teachers or transcription services. 

Our system offers a practical, scalable solution: a full pipeline converting Braille images into natural Chinese text using state-of-the-art large language models fine-tuned with curriculum learning.

In addition to serving educational needs, the system holds long-term potential for integration into national assistive education policies and could catalyze broader adoption of machine learning tools in disability support services. By translating Braille into standard Chinese, the system also empowers visually impaired individuals to engage in digital communication, document sharing, and formal examination settings without requiring costly human transcription. This not only supports educational inclusion but upholds principles of linguistic dignity, cultural equality, and social justice for an often-overlooked population.
\section{Qualitative Case Study on Test Examples}
\label{sec:appendix_case_study}

\begin{figure*}[t]
\centering
    \includegraphics[width=\textwidth]{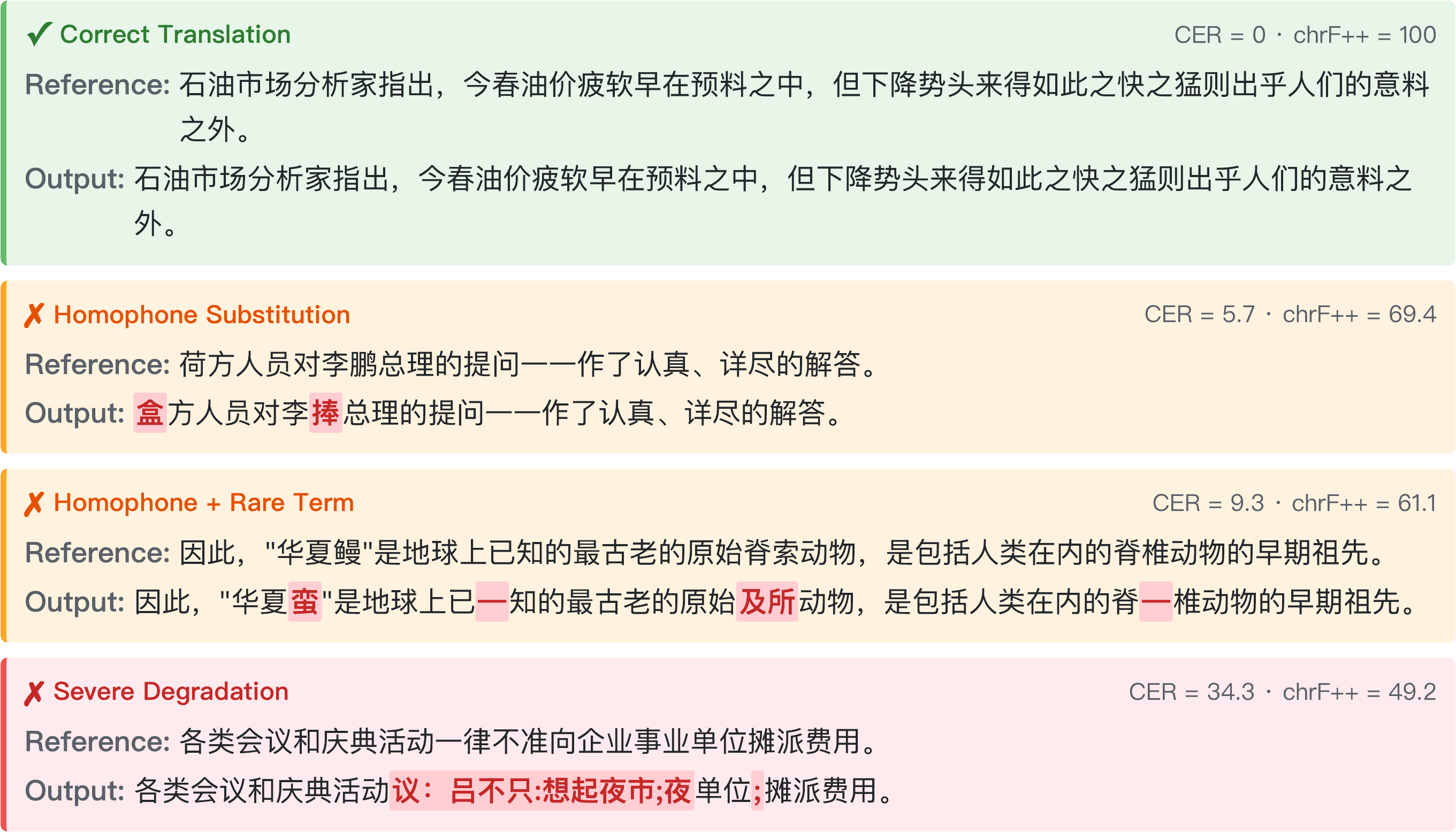}
\caption{Representative translation examples from the passage-level test set (10\% tone retention). Errors are highlighted in \textcolor{red}{red}. The first case shows a successful translation; the remaining cases illustrate increasingly severe failure modes.}
\label{fig:case_study}
\end{figure*}

To qualitatively illustrate the behavior of \methodname, Figure~\ref{fig:case_study} presents four representative translation examples drawn from our held-out passage-level test set under the most aggressive tone-omission setting (10\% tone retention, i.e., 90\% of tone markers removed). These examples come from the same in-distribution test split used in Tables~\ref{tab:main_results} and~\ref{tab:ablation} and are distinct from the real-world expert-transcribed passages evaluated in \S\ref{sec:real_world}; they therefore characterize the behavior underlying our reported test-set metrics rather than deployment performance.

We select one example that the model translates correctly and three that expose distinct failure modes, ordered by increasing severity. Together they show why aggregate metrics (BLEU-4, chrF++, CER) capture only part of the translation quality picture: most remaining errors are phonetically plausible but semantically incorrect, while a small fraction of inputs trigger sustained decoder drift that disproportionately harms character-level metrics.

\paragraph{Successful translation.}
The first example demonstrates that \methodname\ can correctly translate a long sentence containing domain-specific vocabulary (\zh{石油市场}, ``oil market''; \zh{疲软}, ``sluggish'') even when 90\% of tone markers are absent. The model leverages contextual cues to resolve the extensive tonal ambiguity inherent in Braille text. Outputs of this quality account for the bulk of the passage-level test set and underlie the 83.28 BLEU-4 reported in Table~\ref{tab:main_results}.

\paragraph{Homophone confusion.}
The second and third examples illustrate the most common failure mode: homophone substitution. In the second example, the proper noun \zh{李鹏} (L\v{\i} P\'{e}ng) is rendered as \zh{李捧} (L\v{\i} P\v{e}ng), which is phonetically similar but semantically incorrect. Without tone markers, the model cannot distinguish characters that share the same consonant-vowel structure, and for low-frequency proper nouns the surrounding context often does not provide enough signal to recover the correct tone. The third example shows the same pattern on a rare domain term: \zh{华夏鳗} (a fossil species name) is mistranslated as \zh{华夏蛮}, substituting a high-frequency character for the domain-specific one. Errors of this type are precisely what the LLM polishing step (\S\ref{sec:polishing_results}) is designed to correct: the pinyin structure is preserved, so a general-purpose LLM with broader world knowledge can recover the intended character. This explains why polishing yields large BLEU and CER gains (Tables~\ref{tab:polishing_sent}--\ref{tab:polishing_passage}) despite making no change to the underlying Braille model.

\paragraph{Severe degradation.}
The fourth example shows a case where the model loses coherence mid-sentence, producing nonsensical character sequences (\zh{吕不只:想起夜市}). This typically occurs when the input contains a dense cluster of tone-ambiguous syllables with few contextual anchors, causing the decoder to drift into an incoherent state from which it cannot recover within the output budget. Such failures are rare but contribute disproportionately to CER because a single drift event corrupts many downstream characters. They are also the hardest case for the polishing stage, since the polisher operates on the translation alone (without access to the original Braille input) and has no plausible character string to anchor its correction.

\end{document}